\documentclass[conference]{IEEEtran}
\usepackage{cite}
\usepackage{amsmath,amssymb,amsfonts}
\usepackage{algorithmic}
\usepackage{graphicx}
\usepackage{textcomp}
\usepackage{xcolor}

\usepackage{booktabs}    % \toprule, \midrule, \bottomrule
\usepackage{url}
\usepackage[hidelinks]{hyperref}   % no colored boxes; just hyperlinks

\usepackage{soul}
\sethlcolor{white}
\newcommand{\rev}[1]{\hl{#1}}

\graphicspath{{figures/}}

\newcommand{\dataset}{RadioML 2016.10a}
\newcommand{\methodname}{FuseAMC}

\newcommand{\vect}[1]{\mathbf{#1}}
\newcommand{\mat}[1]{\mathbf{#1}}

\def\BibTeX{{\rm B\kern-.05em{\sc i\kern-.025em b}\kern-.08em
    T\kern-.1667em\lower.7ex\hbox{E}\kern-.125emX}}

\newcommand{\authcell}[5]{%
  \parbox[t]{0.30\textwidth}{\centering
    #1\\[0.2ex]
    \textit{#2}\\ \textit{#3}\\ #4\\ #5}%
}

\begin{document}

% -----------------------------------------------------------------------------
% TITLE BLOCK
% -----------------------------------------------------------------------------
\title{Deep Meets Classical: A CPU-Efficient Decision-Fusion Model for Automatic Modulation Classification}

\author{%
\begin{tabular}{@{}c@{\hspace{0.03\textwidth}}c@{\hspace{0.03\textwidth}}c@{}}
\authcell{Zhiyi Yin\textsuperscript{*}}{Department of Computer Science}{Rice University}{Houston, United States}{z.yin@alumni.rice.edu\textsuperscript{*}} &
\authcell{Zhizhao Wen}{Department of Computer Science}{Rice University}{Houston, United States}{zhizhaowen@alumni.rice.edu} &
\authcell{Ruoxin Zhang}{Department of Computer Science}{Rice University}{Houston, United States}{lz37@alumni.rice.edu} \\ \noalign{\vskip 5mm}
\authcell{Chao Wang}{Department of Computer Science}{Rice University}{Houston, United States}{zylj2020@outlook.com} &
\authcell{Puyang Xu}{Department of Electrical and Computer Engineering}{Duke University}{Durham, United States}{rivocxu@gmail.com} &
\authcell{Yifan Jiang}{Department of Electrical and Computer Engineering}{Duke University}{Durham, United States}{yifan.jiang999@gmail.com} \\ \noalign{\vskip 5mm}
\authcell{Yusi Gong}{Department of Computer Science}{Rice University}{Houston, United States}{gyusi73199@gmail.com} &
&
\\
\end{tabular}%
}

\maketitle

% -----------------------------------------------------------------------------
% ABSTRACT
% -----------------------------------------------------------------------------
\begin{abstract}
Automatic modulation classification (AMC) underpins spectrum sensing and
non-cooperative signal identification in cognitive radio, yet the two
standard CPU-feasible approaches to it, a compact convolutional neural
network (CNN) on raw in-phase and quadrature (I/Q) samples and a classical
higher-order-cumulant feature classifier, are rarely compared against each
other conditioned on signal-to-noise ratio (SNR). We show that they are, in
fact, complementary: the CNN is markedly more robust at low SNR, where
hand-crafted statistics collapse toward chance, while the cumulant classifier
is more accurate at high SNR, where higher-order statistics cleanly separate
dense constellations. We exploit this complementarity with \methodname{}, a
lightweight decision-fusion model that trains a logistic-regression
meta-combiner over the concatenated class-posteriors of an approximately
198k-parameter residual CNN and a gradient-boosted-tree cumulant classifier,
on a validation split disjoint from test. On \dataset{} (5 seeds, 44k-frame
held-out test set), \methodname{} reaches 55.8\% overall accuracy,
improving on the best cumulant-feature baseline by 5.1 points and on the CNN
alone by 3.2 points (paired-bootstrap 95\% confidence-interval lower
bounds of 5.0 and 3.1 points, both positive), while matching the best
classical baseline in its own strong high-SNR regime and the CNN in its
strong low-SNR regime. \rev{With about 0.2 million parameters, a model footprint of a few megabytes,
and roughly 3 MFLOPs per frame, the whole pipeline trains and runs on one CPU
core, targeting general-purpose and embedded processors without hardware
acceleration.}
\end{abstract}


% -----------------------------------------------------------------------------
% I. INTRODUCTION
% -----------------------------------------------------------------------------
\section{Introduction}
\label{sec:introduction}

Automatic modulation classification (AMC) is a core task in non-cooperative
spectrum sensing, cognitive radio, and electronic support: a receiver must
identify the modulation scheme of an intercepted waveform before
demodulation, often without side information about the transmitter. Reliable
AMC is a building block of dynamic spectrum access, interference
identification, and physical-layer security in modern communications and
computer engineering systems. Because AMC front-ends are frequently deployed
on embedded software-defined radios and sensor nodes without GPU
accelerators, a CPU-only, laptop-scale evaluation is itself an operationally
relevant regime, not merely a convenience.

Two well-established CPU-feasible approaches address AMC from opposite ends.
The classical approach reduces each waveform to hand-crafted higher-order
cumulants and instantaneous amplitude/phase/frequency statistics, then
classifies with a conventional model such as an SVM or gradient-boosted
tree~\cite{swami2000hierarchical,oshea2018otair}. The deep-learning approach
instead trains a compact CNN directly on the raw I/Q samples, learning its own
features~\cite{oshea2016convolutional}. These lines are almost always compared
only in aggregate, as if one must simply beat the other. We instead ask:
across the SNR range, \emph{where} does each win, and can this be exploited? We
find a clear answer: a compact CNN is far more robust at low SNR, where
hand-crafted cumulant estimates are noisy and collapse toward chance, while the
cumulant classifier is more accurate at high SNR, where higher-order statistics
cleanly separate constellations a plain CNN's first-order pooling does not fully
resolve. Neither structure dominates the other.

This complementarity motivates \methodname{}, a decision-level fusion model: a
logistic-regression meta-combiner over the concatenated class-posteriors of the
CNN and the cumulant gradient-boosted-tree classifier, fit on a validation split
disjoint from test. It is cheap to train, interpretable, and implicitly learns
an SNR-adaptive trust allocation between the streams without seeing an SNR label
at inference.

\noindent The contributions of this work are:
\begin{itemize}
  \item \rev{A model that runs end to end on one CPU core, no GPU: about
  0.2 million parameters, a few megabytes on disk, near 3 MFLOPs per frame, and
  about 2.5 minutes to train per seed} (Sec.~\ref{sec:cost}).
  \item We show, across 5 seeds on \dataset{}, that a compact raw-I/Q CNN and a
  tuned cumulant-feature classifier are complementary across SNR (CNN dominant at
  low SNR, cumulant dominant at high SNR) rather than one uniformly better.
  \item We propose \methodname{} and show it beats \emph{both} pure baselines
  overall and on macro-F1, with a fairly tuned classical baseline (full
  hyperparameter search) so the gain is attributable to fusing complementary
  structures, not a weak baseline or a fancier single network.
\end{itemize}

% -----------------------------------------------------------------------------
% II. RELATED WORK
% -----------------------------------------------------------------------------
\section{Related Work}
\label{sec:related}

\textbf{Deep learning for AMC.} \cite{oshea2016convolutional} introduced
\dataset{} and showed a shallow CNN on raw I/Q (VT-CNN2) beats cumulant-feature
classifiers, especially at low SNR. \cite{west2017deep} finds accuracy is not
limited by network depth on this benchmark, and \cite{oshea2018otair} extends
deep AMC to over-the-air signals with a ResNet baselined against
higher-order-moments plus boosted trees, which validates our classic-baseline
choice. \cite{hou2023clst} distil a conv-transformer (ClST) from a larger
teacher. These reach higher absolute accuracy but need larger compute budgets or
a distillation teacher than the CPU-only regime we target.

\textbf{Classical cumulant-based AMC.} \cite{swami2000hierarchical} established
the canonical hierarchical fourth-order-cumulant classifier
($C_{40}, C_{41}, C_{42}$, normalized by $C_{21}$). \cite{dgani2024cumulant}
revisit it with modern classifiers, showing it stays a competitive, CPU-cheap
alternative to deep models; gradient-boosted trees remain accurate and
CPU-efficient on such low-dimensional engineered
features~\cite{wang2025gpu,zhang2025rag}, motivating keeping a tuned tree-based
classifier as one fusion stream.

\methodname{} sits in the CPU-only, lightweight regime and does not compete with
GPU-scale ResNet/CLDNN ensembles~\cite{west2017deep,oshea2018otair} or
knowledge-distilled conv-transformer hybrids~\cite{hou2023clst}; within a budget
where both a compact CNN and a cumulant classifier train on a CPU in minutes, we
ask whether combining their decisions beats either alone.

% -----------------------------------------------------------------------------
% III. METHOD
% -----------------------------------------------------------------------------
\section{Method}
\label{sec:method}

\subsection{Problem Formulation}
\label{sec:formulation}
Each input is a raw complex baseband I/Q frame $\vect{x} \in
\mathbb{R}^{2\times128}$, consisting of 128 in-phase and 128 quadrature
samples, drawn from one of $|\mathcal{Y}|=11$ modulation classes
$y\in\mathcal{Y}$ (BPSK, QPSK, 8PSK, QAM16, QAM64, GFSK, CPFSK, PAM4, WBFM,
AM-DSB, AM-SSB, as defined by \dataset{}). Each frame is normalized to unit
$L_2$ norm across both channels before being passed to any model or feature
extractor, so that every method (proposed and baselines alike) operates on
directly comparable, power-normalized inputs, matching standard practice in
the AMC literature. The task is standard multi-class classification: produce
a posterior over $\mathcal{Y}$ for each frame, minimizing cross-entropy loss
on the training set.

\subsection{Two Complementary Streams}
\label{sec:streams}

\textbf{CNN stream (raw I/Q).} A compact residual convolutional network
($\sim$198k parameters) is trained end-to-end on the raw $2\times128$ I/Q
frame~\cite{oshea2016convolutional}. Two stride-2 convolutional blocks
(convolution $\to$ batch normalization $\to$ GELU) reduce the temporal
resolution while expanding the channel dimension, followed by a residual
block~\cite{he2016deep}; the resulting feature map is first-order (mean)
pooled and passed through a small MLP head to produce an 11-way softmax
posterior $\vect{p}_{\text{cnn}}(\vect{x}) \in \Delta^{10}$. The CNN is
trained from scratch with the AdamW optimizer~\cite{kingma2015adam,loshchilov2019decoupled} (learning
rate $1.2\times10^{-3}$, weight decay $10^{-4}$, batch size 1024, 12 epochs
with cosine decay and early stopping), with no pretraining or distillation.

\textbf{Cumulant-feature stream.} For the same frame, a 16-dimensional
hand-crafted feature vector is computed: normalized second-, fourth-, and
sixth-order cumulants ($C_{20}, C_{21}, C_{40}, C_{41}, C_{42}, C_{60},
C_{63}$) together with instantaneous amplitude, phase, and frequency
statistics~\cite{swami2000hierarchical,oshea2018otair}. A gradient-boosted
tree classifier, tuned by random search (Sec.~\ref{sec:setup}), maps this
feature vector to an 11-way posterior $\vect{p}_{\text{gbt}}(\vect{x}) \in
\Delta^{10}$.

The two streams are trained independently and never share parameters or
gradients; Sec.~\ref{sec:motivation} shows they err in largely non-overlapping
SNR regimes.

\subsection{Decision Fusion}
\label{sec:fusion}
\begin{figure}[t]
  \centering
  \includegraphics[width=\columnwidth]{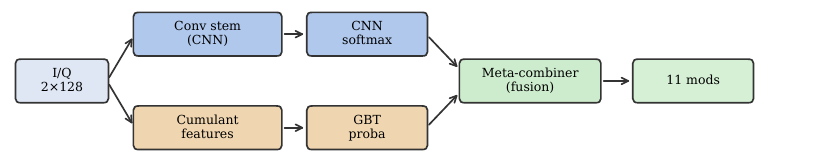}
  \caption{\methodname{} architecture: independent CNN (raw I/Q) and
  cumulant-feature+GBT streams each produce an 11-way posterior; a
  logistic-regression meta-combiner, trained on the validation split, fuses
  the concatenated posteriors into the final prediction.}
  \label{fig:arch}
\end{figure}

\methodname{} fuses the two streams at the decision level
(Fig.~\ref{fig:arch}): the posteriors are concatenated into
$\vect{z}(\vect{x}) = [\vect{p}_{\text{cnn}}(\vect{x}) \,\Vert\,
\vect{p}_{\text{gbt}}(\vect{x})] \in \mathbb{R}^{22}$ and a multinomial
logistic regression with $\mat{W}\in\mathbb{R}^{11\times22}$, $\vect{b}\in
\mathbb{R}^{11}$ produces
\begin{equation}
\vect{p}_{\text{fuse}}(\vect{x}) = \mathrm{softmax}\big(\mat{W}\,\vect{z}(\vect{x}) + \vect{b}\big).
\label{eq:fusion}
\end{equation}
$\mat{W}$ and $\vect{b}$ are fit on the \emph{validation} split, disjoint from
the streams' training data and from test, and the $11\times22$ parameter count
makes overfitting unlikely. Because $\mat{W}$ is learned rather than fixed, the
combiner up-weights whichever stream is more reliable in a region of posterior
space (implicitly, an SNR regime) without an SNR label at inference; zeroing one
stream's block of $\mat{W}$ recovers the other stream alone
(Sec.~\ref{sec:ablation}).

% -----------------------------------------------------------------------------
% IV. EXPERIMENTS
% -----------------------------------------------------------------------------
\section{Experiments}
\label{sec:experiments}

\subsection{Experimental Setup}
\label{sec:setup}

\textbf{Dataset and splits.} We use \dataset{}
\cite{oshea2016convolutional,oshea2016dataset}, a synthetic
GNU-Radio-generated dataset of 11 modulation schemes at 20 SNR levels from
$-20$ to $+18$\,dB in 2\,dB steps, with 1000 frames per (modulation, SNR)
cell. For CPU-budget reasons we use a fixed stratified subset: 100 training /
50 validation / 200 test frames per (modulation, SNR) cell, giving
approximately 22k training, 11k validation, and 44k test frames. The split is
drawn once with a fixed split seed, so the 44k-frame test set is identical
across every method and every seed, enabling paired comparisons; the
validation split used to fit the fusion meta-combiner (Sec.~\ref{sec:fusion})
is likewise identical across seeds and disjoint from test. Per-frame $L_2$
normalization is applied identically before any feature extraction or model
input (Sec.~\ref{sec:formulation}).

\textbf{Training and tuning.} The CNN stream is trained with a single fixed
recipe (learning rate, weight decay, batch size, epoch budget, cosine
schedule, early stopping) identical across all five reporting seeds
$\{0,1,2,3,4\}$; it is \emph{not} hyperparameter-tuned. The cumulant-feature
classifiers (GBT, random forest, SVM) instead each receive a random search
(12 trials) over their respective hyperparameter spaces on a tuning seed
disjoint from the reporting seeds. This asymmetry is deliberately conservative
for our own method: the classical baselines are given the tuning advantage, so
any margin \methodname{} achieves over them cannot be attributed to an
undertuned baseline. The fusion meta-combiner (Eq.~\eqref{eq:fusion}) is
fit with standard L2-regularized logistic regression on the validation split
for each seed, using that seed's already-trained CNN and GBT.

We pre-committed, before the final experiment, to two numeric criteria (both
met, Sec.~\ref{sec:results}): \methodname{} must exceed the best
cumulant-feature baseline by $\geq 0.05$ overall accuracy and the CNN alone by
$\geq 0.02$, each with a paired-bootstrap 95\% CI lower bound above zero.

\textbf{Metrics and statistical testing.} We report overall test accuracy,
low-SNR accuracy (SNR $\leq 0$\,dB), high-SNR accuracy (SNR $\geq 6$\,dB), and
macro-F1, each as mean $\pm$ standard deviation over the five reporting
seeds. To assess whether an accuracy gap is reliable rather than a
seed-variance artifact, we compute the paired-bootstrap 95\% confidence
interval of the accuracy difference across seeds, and treat an improvement as
established only when the CI lower bound is greater than zero.

\subsection{Motivating Observation: Complementary Errors Across SNR}
\label{sec:motivation}
The premise behind \methodname{} is that the two streams fail in different SNR
regimes. At low SNR ($\leq 0$\,dB) the CNN reaches $33.6\%$ versus $24.1\%$ for
the best cumulant baseline; at high SNR ($\geq 6$\,dB) the ordering reverses
($84.3\%$ vs.\ $76.2\%$). Each stream is stronger by $8$ to $9$ points in the
opposite half of the range, so fusing rather than choosing between them is the
natural response.

\subsection{Main Results}
\label{sec:results}

Table~\ref{tab:main_results} reports overall, low-SNR, and high-SNR accuracy
(mean $\pm$ std over 5 seeds) for \methodname{} against the canonical
cumulant-feature baselines (SVM, random forest, gradient-boosted trees) and
the CNN-alone reference, together with macro-F1. Paired-bootstrap 95\% CIs of
the accuracy gap against each baseline are reported inline below and, for the
two fusion streams, in Table~\ref{tab:ablation}.

\begin{table}[t]
\caption{Test accuracy by SNR regime (Low: SNR$\leq$0\,dB; High: SNR$\geq$6\,dB) and macro-F1 (M-F1) on RadioML 2016.10a (44k frames), mean $\pm$ std over 5 seeds. Best per column in bold.}
\label{tab:main_results}
\centering
\setlength{\tabcolsep}{3pt}
\resizebox{\columnwidth}{!}{%
\begin{tabular}{lcccc}
\toprule
Method & Overall & Low & High & M-F1 \\
\midrule
Cumulant+SVM & $0.507 \pm 0.001$ & $0.241 \pm 0.001$ & \textbf{$0.843 \pm 0.001$} & $0.525 \pm 0.001$ \\
Cumulant+RF & $0.498 \pm 0.000$ & $0.237 \pm 0.001$ & $0.830 \pm 0.001$ & $0.517 \pm 0.000$ \\
Cumulant+GBT & $0.496 \pm 0.001$ & $0.236 \pm 0.001$ & $0.827 \pm 0.002$ & $0.512 \pm 0.002$ \\
CNN (raw I/Q) & $0.526 \pm 0.002$ & $0.336 \pm 0.002$ & $0.762 \pm 0.003$ & $0.526 \pm 0.003$ \\
\midrule
Avg. fusion & $0.557 \pm 0.001$ & $0.332 \pm 0.002$ & $0.837 \pm 0.002$ & $0.572 \pm 0.001$ \\
Fusion (ours) & \textbf{$0.558 \pm 0.001$} & \textbf{$0.337 \pm 0.002$} & $0.835 \pm 0.001$ & \textbf{$0.575 \pm 0.001$} \\
\bottomrule
\end{tabular}}
\end{table}

\begin{figure}[t]
  \centering
  \includegraphics[width=\columnwidth]{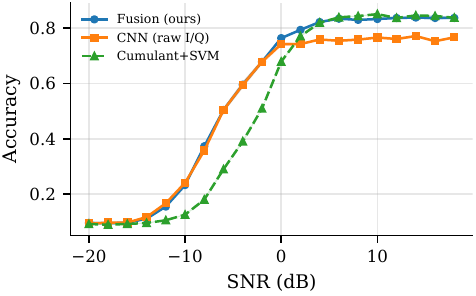}
  \caption{Test accuracy vs.\ SNR for \methodname{}, the CNN alone, and the
  best cumulant-feature baseline; \rev{lines are the 5-seed mean, shaded bands
  are plus/minus one standard deviation}. The CNN and cumulant curves cross over near
  $0$\,dB (the CNN is higher below it, the cumulant classifier higher above
  it), and \methodname{} follows the upper envelope of both across the range.}
  \label{fig:acc_vs_snr}
\end{figure}

\methodname{} reaches $55.8\%$ overall accuracy, an improvement of
$+5.1$ points over the strongest cumulant-feature baseline (SVM; 95\% CI
$[+5.0, +5.2]$) and $+3.2$ points over the CNN alone (95\% CI $[+3.1, +3.4]$),
meeting both pre-committed win criteria (Table~\ref{tab:main_results}); the
aggregate mean and CI lower bound clear both margins, and all five seeds
individually clear the CNN margin (four of five clear the cumulant margin, the
fifth at $+0.0497$). It also attains the best macro-F1 ($0.575$ vs.\ $0.526$). Figure~\ref{fig:acc_vs_snr} shows why:
\methodname{}
tracks the CNN at low SNR (where it reaches $33.7\%$, versus $24.1\%$ for the
best cumulant baseline) and tracks the cumulant baseline at high SNR (reaching
$83.5\%$, $0.8$ points below the best classical model and $+7.3$ points
above the CNN), following the upper envelope of the two streams. \rev{At every
SNR it stays within 0.023 of the better stream, exceeding both by up to about 2
points in the 0 to 2 dB crossover; the per-SNR seed spread is under 0.013.} As
expected, all methods converge toward near-chance accuracy ($\sim 1/11$) below
about $-10$\,dB, where no method can recover the signal.

Even at $+18$\,dB the residual errors concentrate in two physically expected
pairs: QAM16 versus QAM64 (nested dense constellations differing only in
their outer points) and the analog WBFM versus AM-DSB (hard to separate during
low-modulation segments); \methodname{} inherits these intrinsic
single-stream confusions rather than removing them.

\subsection{Ablation and Alternative Fusion Rules}
\label{sec:ablation}

Table~\ref{tab:ablation} isolates each stream by removing it from the fusion
(the CNN-alone and GBT-alone rows are the degenerate fusion with one posterior
block zeroed, Sec.~\ref{sec:fusion}). Dropping the CNN lowers low-SNR accuracy
from $33.7\%$ to $23.6\%$ ($-10.1$ points); dropping the GBT lowers high-SNR
accuracy from $83.5\%$ to $76.2\%$ ($-7.3$ points). Overall the GBT and CNN
streams are $6.2$ and $3.2$ points below \methodname{} (both 95\% CI lower
bounds $>0$).

\begin{table}[t]
\caption{Fusion improvement in overall accuracy over each of its two streams (CNN, GBT) and over the naive posterior-average control, with 95\% paired-bootstrap CIs over 5 seeds. Removing either stream, or replacing the learned combiner with a fixed average, strictly reduces accuracy.}
\label{tab:ablation}
\centering
\setlength{\tabcolsep}{4pt}
\begin{tabular}{lcc}
\toprule
Comparison & $\Delta$ Overall Acc. & 95\% CI \\
\midrule
Fusion $-$ CNN stream & $+0.032$ & $[+0.031, +0.034]$ \\
Fusion $-$ GBT stream & $+0.062$ & $[+0.061, +0.063]$ \\
Fusion $-$ Avg.\ fusion & $+0.002$ & $[+0.001, +0.003]$ \\
\bottomrule
\end{tabular}
\end{table}

Table~\ref{tab:fusion_variants} \rev{asks whether the gain needs a learned
combiner. Over the same frozen posteriors, every soft rule (posterior averaging,
validation-weighted averaging, temperature-calibrated averaging, a
one-hidden-layer MLP) lands within 0.4 points overall; the learned logistic
combiner is best but only 0.2 points over plain averaging. SNR-dependent hard
switching is worse (0.551 with an oracle SNR threshold, 0.555 with a learned
SNR-free router) because it discards the losing stream instead of down-weighting
it. The gain thus comes from combining the two posteriors at all; the learned
combiner is a small, reliable refinement.}

\begin{table}[t]
\caption{Decision-fusion rules over the same frozen per-seed CNN and GBT posteriors, mean $\pm$ std over 5 seeds. Weighted averaging tunes its weight on validation ($\bar\alpha=0.57$); the oracle switch picks the SNR threshold on validation from the true SNR label. Every soft rule is within $0.4$ points overall. Best in bold.}
\label{tab:fusion_variants}
\centering
\setlength{\tabcolsep}{4pt}
\begin{tabular}{lcc}
\toprule
Fusion rule & Overall acc. & Macro-F1 \\
\midrule
CNN stream & $0.527 \pm 0.002$ & $0.527 \pm 0.004$ \\
GBT stream & $0.496 \pm 0.001$ & $0.512 \pm 0.002$ \\
\midrule
Posterior averaging & $0.556 \pm 0.001$ & $0.572 \pm 0.002$ \\
Weighted averaging & $0.557 \pm 0.001$ & $0.571 \pm 0.002$ \\
Temp.-calibrated avg. & $0.556 \pm 0.001$ & $0.572 \pm 0.002$ \\
SNR hard switch, oracle & $0.551 \pm 0.002$ & $0.562 \pm 0.003$ \\
SNR hard switch, learned & $0.555 \pm 0.002$ & $0.568 \pm 0.003$ \\
Nonlinear meta (MLP) & $0.555 \pm 0.003$ & $0.568 \pm 0.011$ \\
Temp.-calibrated logistic & $0.558 \pm 0.001$ & $0.575 \pm 0.001$ \\
\midrule
Logistic combiner (ours) & \textbf{$0.559 \pm 0.002$} & $0.575 \pm 0.002$ \\
\bottomrule
\end{tabular}
\end{table}

Table~\ref{tab:calibration} \rev{reports each stream's expected calibration
error (ECE). Aggregate ECE is under 2 percent for both, but rises to 6 to 8
percent within the low- and high-SNR regimes, where opposite-signed bins no
longer cancel; the fitted temperatures are 1.06 (CNN) and 0.95 (GBT). Applying
temperature or isotonic calibration to the streams changes fused accuracy by at
most 0.0004: the combiner's validation-fit weights already subsume per-stream,
per-class scaling.}

\begin{table}[t]
\caption{Expected calibration error (ECE, \%, 15 bins) of each stream's posteriors, raw and after validation-fit temperature or isotonic calibration, over 5 seeds. Aggregate ECE is low for both streams, but rises to 6--8\% within each SNR regime; calibrating the streams barely moves the learned combiner.}
\label{tab:calibration}
\centering
\setlength{\tabcolsep}{4pt}
\begin{tabular}{llccc}
\toprule
Stream & Calibration & Overall & Low SNR & High SNR \\
\midrule
CNN & none & 1.6 & 6.6 & 7.3 \\
 & temperature ($T{=}1.06$) & 1.5 & 5.6 & 7.8 \\
 & isotonic & 1.1 & 5.7 & 6.5 \\
\midrule
GBT & none & 1.2 & 5.9 & 7.6 \\
 & temperature ($T{=}0.95$) & 1.3 & 6.7 & 7.2 \\
 & isotonic & 1.2 & 5.6 & 6.9 \\
\bottomrule
\end{tabular}
\end{table}

\subsection{Computational Cost}
\label{sec:cost}

\rev{All execution is single-core with no GPU. The CNN stream dominates
arithmetic cost: 197,867 parameters (0.79 MB in single precision) and about 1.6
million multiply-accumulates per frame (roughly 3.3 MFLOPs, analytical from
layer dimensions). The gradient-boosted tree is the largest artifact on disk:
up to 400 rounds, one shallow tree per class per round ($\sim$4,000 trees,
depth $\leq 3$), about 4 MB pickled or roughly 1.2 MB compressed; the
meta-combiner (253 parameters) is a few KB. The full pipeline is about 0.2
million parameters, roughly 5 MB uncompressed (about 2 MB compressed), with
arithmetic within two percent of the CNN alone, and trains end to end, both
streams and the combiner, in 150 to 160 seconds per seed on one modern CPU
core, within the storage and compute budget of embedded software-defined-radio
processors.}

\subsection{Discussion}
\label{sec:discussion}

\methodname{} meets both pre-committed criteria: it beats the best
cumulant-feature baseline by $+5.1$ points overall (95\% CI $[+5.0,+5.2]$) and
the CNN alone by $+3.2$ points overall (95\% CI $[+3.1,+3.4]$), and
Sec.~\ref{sec:motivation} confirms it does so by tracking whichever stream is
better in a given SNR regime rather than by a uniform improvement. We are
careful not to overclaim: at high SNR the SVM cumulant baseline is marginally
higher than \methodname{} ($84.3\%$ vs.\ $83.5\%$), so fusion
\emph{matches}, rather than strictly beats, the best classical model there; its
advantage is overall, on macro-F1, and at low SNR. The main caveat is scale:
our reduced stratified split ($\sim$22k training frames) keeps absolute
accuracies below full-dataset published numbers, though the identical split
and test set keep paired comparisons valid. Earlier
single-network variants (temporal self-attention, second-order covariance
pooling) did not beat the plain CNN here~\cite{wen2025ssd,wang2025gpu}: the gain
comes from fusing two structurally different predictors, not a fancier network.

Table~\ref{tab:impairments} \rev{stress-tests the clean-trained models under
test-time shifts absent from training. A sample-rate offset, 2-tap fading, extra
noise, or a residual carrier-frequency offset lowers fused accuracy by 0.08 to
0.24, with carrier offset the most damaging. Fusion tracks the two streams down
under every shift rather than repairing it, and under fading a fixed posterior
average (the same rule as the Avg. fusion row of Table}~\ref{tab:main_results}%
\rev{, reaching 0.505 here) is more robust than the tabulated combiner (0.470),
whose weights over-trust the CNN once the channel changes. A full cross-dataset study on
RadioML~2018.01A and over-the-air captures is the primary next step.}

\begin{table}[t]
\caption{Robustness to test-time channel/hardware shift absent from training: overall accuracy (mean over 5 seeds) when the 44k-frame test set is perturbed; models are trained only on the clean RadioML~2016.10a split. $\Delta$ is the fusion change from its clean accuracy.}
\label{tab:impairments}
\centering
\setlength{\tabcolsep}{4pt}
\begin{tabular}{lcccc}
\toprule
Test-time shift & CNN & GBT & Fusion & $\Delta$Fus. \\
\midrule
clean (reference) & 0.527 & 0.496 & 0.559 & --- \\
\midrule
sample-rate offset $+1\%$ & 0.450 & 0.460 & 0.482 & $-0.077$ \\
2-tap fading, $|a|{=}0.3$ & 0.409 & 0.469 & 0.470 & $-0.089$ \\
added noise, $-6$\,dBc & 0.448 & 0.398 & 0.464 & $-0.095$ \\
added noise, $-3$\,dBc & 0.351 & 0.325 & 0.361 & $-0.197$ \\
carrier offset $0.005$/sample & 0.364 & 0.312 & 0.351 & $-0.208$ \\
carrier offset $0.01$/sample & 0.310 & 0.310 & 0.321 & $-0.238$ \\
\bottomrule
\end{tabular}
\end{table}

% -----------------------------------------------------------------------------
% V. CONCLUSION
% -----------------------------------------------------------------------------
\section{Conclusion}
\label{sec:conclusion}

We showed that a compact raw-I/Q CNN and a tuned cumulant-feature classifier
are complementary across SNR on \dataset{} (the CNN more robust at low SNR,
the cumulant classifier more accurate at high SNR), and exploited this with
\methodname{}, a lightweight, interpretable decision-fusion model that trains
a logistic-regression meta-combiner over the two streams' posteriors on a
held-out validation split. Across 5 seeds, \methodname{} beats both pure
baselines overall ($+5.1$ points over the best cumulant-feature baseline,
$+3.2$ points over the CNN alone, both with 95\% CI lower bound $>0$) and on
macro-F1, dominating at low SNR and matching the best classical model at high
SNR, entirely on a single CPU core \rev{(about 0.2 million parameters, roughly
5 MB uncompressed on disk (2 MB compressed), near 3.3 MFLOPs per frame), within
the budget of the embedded processors in software-defined radios; compressing
the gradient-boosted tree and quantizing the CNN are concrete next steps}. Its
clearest limitation is scale: results
are on a reduced stratified split for CPU tractability, and validating the
fusion principle at full dataset scale is left to future work, along with
extending the
fusion principle to additional complementary feature families and evaluating
on the larger RadioML 2018.01A benchmark.

% -----------------------------------------------------------------------------
% CODE AND DATA AVAILABILITY
% -----------------------------------------------------------------------------
\section{Code and Data Availability}
To support reproducibility, our code, the fixed stratified train/validation/test
splits, and the per-seed run logs underlying every reported number are available
at \url{https://github.com/crayzy519/fuseamc} and will be made public upon
publication. \dataset{} is itself a public benchmark.

% -----------------------------------------------------------------------------
% REFERENCES
% -----------------------------------------------------------------------------
\bibliographystyle{IEEEtran}
\bibliography{references}

\end{document}